%% file: acl_latex.tex
\documentclass[11pt]{article}

\usepackage[preprint]{acl}

\usepackage{times}
\usepackage{latexsym}

\usepackage[T1]{fontenc}

\usepackage{CJKutf8}
\usepackage[utf8]{inputenc}
\usepackage{algorithm}
\usepackage{algorithmicx}
\usepackage{algpseudocode}

\usepackage{microtype}

\usepackage{inconsolata}

\usepackage{graphicx}
\usepackage{amsmath}
\usepackage{amsfonts}
\usepackage{bbm}

\usepackage{tikz-dependency}
\usepackage{graphicx}
\usepackage{tikz}
\usepackage{tikz-qtree}
\usepackage{caption}
\usepackage{bbm}
\usepackage{letltxmacro}
\usepackage{mathtools}
\usepackage{pgfplots}
\usepackage{dashrule}
\usepackage[hang,flushmargin]{footmisc} %
\usepackage{graphics}
\usepackage{pdfpages}
\usepackage{subcaption} 
\usepackage{xpinyin}
\usetikzlibrary{tikzmark, arrows.meta}
\usepackage{xcolor,color}
\usepackage{algorithm}     %
\usepackage{algorithmicx}  %
\usepackage{algpseudocode}
\usepackage{amsmath}
\usepackage{colortbl}
\usepackage{booktabs}
\usepackage{multirow}
\definecolor{highlight}{RGB}{0,0,0}
\definecolor{revise}{RGB}{0,0,0}

\definecolor{clearpurple}{RGB}{138, 140, 191}
\definecolor{clearyellow}{HTML}{f2d3bf}
\definecolor{skyblue}{HTML}{a8d8e2}
\definecolor{darkblue}{HTML}{19183B}
\definecolor{tp1}{RGB}{253, 207, 158}

\definecolor{brickred}{HTML}{b92622}
\definecolor{midnightblue}{HTML}{005c7f}
\definecolor{salmon}{HTML}{f1958d}
\definecolor{burntorange}{HTML}{f19249}
\definecolor{junglegreen}{HTML}{4dae9d}
\definecolor{forestgreen}{HTML}{499c5e}
\definecolor{pinegreen}{HTML}{3d8a75}
\definecolor{seagreen}{HTML}{6bc1a2}
\definecolor{limegreen}{HTML}{97c65a}
\definecolor{pink}{HTML}{E8A0BF}
\definecolor{purple}{HTML}{7149C6}
\definecolor{mylightgray}{HTML}{E5E1DA}
\definecolor{myorange}{HTML}{FFB266} %
\definecolor{mygreen}{HTML}{82CA9D}  %
\definecolor{myred}{HTML}{F08080}    %

\title{JSPG: Dynamic Dictionary Filtering via Joint Semantic-Pinyin-Glyph Retrieval for Chinese Contextual ASR}

\author{
    Shilin Zhou,
    Zhenghua Li\Thanks{$~$ Corresponding author}\\
    School of Computer Science and Technology, \\
    Soochow University, Suzhou, China \\
    \texttt{slzhou.cs@outlook.com; zhli13@suda.edu.cn} \\
}

\begin{document}
\begin{CJK}{UTF8}{gkai}
\maketitle
\begin{abstract}
Contextual Automatic Speech Recognition (ASR) faces challenges with large-scale keyword dictionaries, as excessive irrelevant candidates introduce noise that degrades accuracy. To address this, dynamic filtering typically uses a base ASR model to generate preliminary hypotheses, followed by semantic text retrievers to fetch a concise subset of relevant keywords. However, this approach frequently fails in Chinese ASR. Base models often produce homophonic or near-homophonic errors that preserve the phonetic cues of the target keywords but severely distort their semantic meaning, rendering standard semantic retrievers ineffective. 
To resolve this, we propose a filtering framework that jointly integrates Semantic, Pinyin, and Glyph features (JSPG).
Pinyin effectively retrieves targets based on phonetic similarity, while glyph provides complementary structural cues to filter out numerous irrelevant homophones inherent in Chinese.
To bridge the gap between character-level pinyin/glyph metrics and sequence-level filtering, we introduce an extended Smith-Waterman algorithm that computes similarity scores between the N-best hypothesis sequences and keywords.
Experiments on the Aishell-1 and RWCS-NER datasets demonstrate that JSPG significantly outperforms single-feature baselines. Furthermore, downstream contextual ASR models guided by JSPG achieve substantial improvements in keyword recognition accuracy.

\end{abstract}

\input{contents/intro.tex}

\input{contents/related_work.tex}

\input{contents/approach.tex}

\input{contents/exp.tex}

\input{contents/conclusion.tex}

\section*{Limitations}
While JSPG demonstrates promising results, we acknowledge two limitations in this work.

First, our evaluation is currently limited to the Chinese language. The specific implementation of our method relies on Chinese pinyin and character structures. However, we believe that the core principle of leveraging phonetic and structural features is universal. With appropriate modifications, we expect our approach can be adapted to other languages that face similar homophonic challenges, such as Japanese or Korean.

Second, this work mainly focuses on the retrieval stage. We aim to find the most relevant keywords to narrow down the search space. 
We have not optimized how the downstream contextual ASR model utilizes these keywords. Since the retrieved candidates are often phonetically or visually similar to the target, guiding the model to strictly distinguish the correct keyword from these confusing candidates is a critical step. 
We plan to explore better ways to utilize the retrieved context in future work.

\bibliography{custom}

\appendix

\section{Appendix}
\label{sec:appendix}
\input{contents/sw_example}

\end{CJK}
\end{document}

%% file: contents/intro.tex
\section{Introduction}

Automatic Speech Recognition (ASR) systems have achieved remarkable success in general transcription tasks \cite{radford-2022-whisper,bai-etal-2024-seedasr, omnilingual2025omnilingual}. However, accurately recognizing contextual keywords, such as technical terminologies and rare entities, remains a challenge. 
The keywords appear infrequently in general training corpora, which makes it difficult for models to transcribe them effectively \cite{peters2018deep,sudo2024contextualized}. 

To address this, researchers have proposed the contextual ASR task \cite{pundak-2018-deep, alon2019contextual, zhou-etal-2024-copyne}, which utilizes a predefined dictionary of potential keywords to guide the recognition process. 
Current approaches have achieved promising results when the keyword dictionary is small.
However, experiments show that as the dictionary size increases, the presence of irrelevant keywords leads to a significant decline in transcription accuracy \cite{zhou-etal-2024-copyne, sudo-2024-contextualized,zhou2025improving}.

\input{figs/example_new}

To alleviate the negative impact of large dictionaries, recent work has explored dynamically filtering the dictionary into a high-relevance subset of keywords for each speech input \cite{ mathur-etal-2024-doc, li-etal-2024-rag, chen-etal-2025-wavrag, lei-etal-2025-contextualization}. 
In a typical cascaded pipeline, a base ASR model first transcribes the speech input into preliminary N-best hypotheses.
Then these methods usually either select the top-ranked hypothesis as the query or concatenate all hypotheses into a single query sequence.
Subsequently, the query and the keywords in the dictionary are fed into a pre-trained text embedding model, which converts them into embedding representations.
Finally, the system calculates the cosine similarity scores between them to select the top-scoring keywords as the filtered subset \cite{ xiao-etal-2025-contextual, dimitrios-etal-2025-retrieval}.
Since text embeddings are designed to encode the contextual meaning of the input, we refer to these methods as semantic retrievers. 

However, relying solely on semantic features is insufficient for precise keyword filtering in Chinese contextual ASR.
Since ASR is acoustic-based and Chinese has abundant homophones, ASR errors typically sound similar to the target keyword but carry entirely different meanings. 
Semantic retrievers fetch keywords that are semantically related to the erroneous transcription but completely irrelevant to the actual target. For instance, as illustrated in Figure \ref{fig:retrieval_process}, the target ``期权'' (Options) is misrecognized as ``弃权'' (Abstention) due to homophonic confusion. Consequently, the retriever incorrectly fetches the irrelevant keyword ``放弃'' (Give Up) based on its high semantic similarity with ``弃权''.

Since these errors preserve phonetic similarity to the target, we can leverage this cue for retrieval. In Chinese, pinyin directly encodes pronunciation, making it a natural metric for phonetic matching. 
However, many different Chinese characters share the exact same pronunciation. Therefore, relying solely on pinyin introduces substantial noise, as it retrieves many irrelevant characters alongside the correct keyword.

Fortunately, the glyph features of Chinese characters offer a complementary solution to this phonetic noise. Since most characters are phono-semantic compounds, ASR phonetic errors frequently share structural components with the target keywords \cite{tan-etal-2005-neuroanatomical}.
Crucially, while many different characters share the exact same pronunciation, every individual character has a unique glyph structure.
This one-to-one property creates sharp discriminative signals, allowing the system to easily distinguish the target from unrelated homophones retrieved by pinyin.

In this work, we propose JSPG, a Joint Semantic-Pinyin-Glyph retrieval approach for Chinese contextual ASR. 
JSPG integrates two types of features: a global \textit{semantic} score derived from pre-trained text embeddings to capture the broad context \cite{mathur-etal-2024-doc, lei-etal-2025-contextualization}, and a unified \textit{phonetic-glyph} score that fuses character-level pinyin and glyph similarities \cite{qiao-etal-2025-disc} to ensure precise local matching. 
Since pinyin and glyph metrics inherently operate at the character level, we introduce an extended Smith-Waterman algorithm \cite{smith-waterman-1981-identification} that extends them to measure the sequence-level similarity between a short keyword and a longer, noisy ASR hypothesis sequence.

We evaluate JSPG on the standard Aishell-1 \cite{bu-2017-aishell, chen-2022-aishell} and the complex, real-world RWCS-NER datasets \cite{zhou-etal-2024-chinese}. Experimental results demonstrate that JSPG significantly outperforms baselines relying solely on individual semantic, pinyin, or glyph features. 
The ablation studies further reveal that each component contributes uniquely to the system, with their integration yielding the most robust retrieval performance. 
Furthermore, when contextual ASR models are guided by the keywords filtered by JSPG, they achieve substantial improvements in keyword recognition accuracy.

%% file: figs/example_new.tex
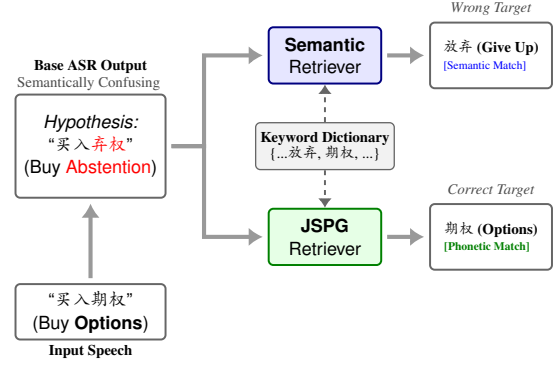
\begin{figure}[!tb]
    \centering
    \scalebox{0.62}{
    \begin{tikzpicture}[
        >=latex, 
        font=\sffamily,
        connect/.style={
                rounded corners=4pt,
                very thick,
                draw=black!80
            },
        fat arrow/.style={
                arrows = {-Latex[length=3mm, width=4mm]},
                shorten >= 3pt,
                shorten <= 3pt,
                line width=2.5pt, 
                draw=black!40
            },
        dict arrow/.style={ 
                arrows = {-Latex[length=2mm, width=3mm]},
                very thick,
                draw=black!60,
                dashed %
            },
        main box/.style={
                rectangle,
                rounded corners=1mm,
                very thick,
                draw=black!60,
                align=center,
                fill=white,
                inner sep=1.5mm
            },
        sem box/.style={
                main box,
                fill=blue!10,
                draw=blue!50!black,
                minimum width=2.4cm,
                minimum height=1.2cm
            },
        pho box/.style={
                main box,
                fill=green!10,
                draw=green!50!black,
                minimum width=2.4cm,
                minimum height=1.2cm
            },
        res box/.style={
                main box,
                minimum width=2.6cm,
                minimum height=1.4cm,
                font=\small,
                align=left
            },
        dict box/.style={
                rectangle,
                rounded corners=1mm,
                thick,
                draw=black!60,
                align=center,
                minimum width=2.8cm,
                minimum height=1.0cm,
                fill=black!5,
                inner sep=1mm,
                font=\footnotesize
            },
        title text/.style={
                align=center,
                anchor=south,
                font=\bfseries\small
            },
        note text/.style={
                font=\footnotesize,
                align=center,
                color=black!70
            }
        ]

        \node[main box, minimum width=3.2cm, minimum height=2.2cm] (node_asr) at (0,0) {
            \textit{Hypothesis:} \\
            ``买入\textcolor{red}{弃权}'' \\
            (Buy \textcolor{red}{Abstention})
        };
        \node[title text] at (node_asr.north) [yshift=3mm] {Base ASR Output};

        \node[note text, anchor=north] at (node_asr.north) [yshift=5mm] {
            Semantically Confusing
        };

        \node[main box, below=1.8cm of node_asr, minimum width=3.2cm] (node_input) {
            ``买入\textbf{期权}'' \\
            (Buy \textbf{Options})
        };
        \node[title text] at (node_input.south) [yshift=-5mm] {Input Speech};
        \draw[fat arrow] (node_input.north) -- (node_asr.south);

        \node[dict box] (dict) at (5.0, 0) {
            \textbf{Keyword Dictionary} \\
            \{...放弃, 期权, ...\}
        };

        \node[sem box, above=0.8cm of dict] (sem_mod) {
            \textbf{Semantic} \\ Retriever
        };

        \node[pho box, below=0.8cm of dict] (pho_mod) {
            \textbf{JSPG} \\ Retriever
        };

        \draw[fat arrow] (node_asr.east) -- ++(0.8,0) |- (sem_mod.west);
        \draw[fat arrow] (node_asr.east) -- ++(0.8,0) |- (pho_mod.west);

        \draw[dict arrow] (dict.north) -- (sem_mod.south);
        \draw[dict arrow] (dict.south) -- (pho_mod.north);

        \node[res box, right=1.0cm of sem_mod] (sem_res) {
            \textbf{放弃 (Give Up)} \\
            \textcolor{blue}{\scriptsize [Semantic Match]}
        };
        \node[note text, above=0mm of sem_res] {\textit{Wrong Target}};

        \node[res box, right=1.0cm of pho_mod] (pho_res) {
            \textbf{期权 (Options)} \\
            \textcolor{green!50!black}{\scriptsize \textbf{[Phonetic Match]}}
        };
        \node[note text, above=0mm of pho_res] {\textit{Correct Target}};

        \draw[fat arrow] (sem_mod.east) -- (sem_res.west);
        \draw[fat arrow] (pho_mod.east) -- (pho_res.west);

    \end{tikzpicture}}
    \caption{
    Illustration of the retrieval process.
    \textbf{Upper Path:} The semantic retriever is misled by the ASR error ``弃权 (Abstention)'' and retrieves ``放弃 (Give Up)''.
    \textbf{Lower Path:} Our JSPG retriever utilizes the joint semantic-pinyin-glyph features to correctly retrieve the target ``期权 (Options)''.
    }
    \label{fig:retrieval_process}
\end{figure}

%% file: contents/related_work.tex
\section{Related Work}

Standard Contextual ASR approaches face significant scalability bottlenecks when handling massive dictionaries, as the influx of irrelevant candidates introduces noise that degrades recognition accuracy \cite{zhou-etal-2024-copyne, sudo-2024-contextualized}.
To address this, recent research adopts dynamic dictionary filtering, which retrieves a concise, high-relevance keyword subset for each speech input before feeding it into the downstream contextual ASR model.
Existing implementations of this mechanism typically fall into two primary categories: text-based cascaded retrieval and audio-based end-to-end retrieval.

\subsection{Text-based Cascaded Retrieval}
The first category, text-based cascaded retrieval, operates by first utilizing a lightweight base ASR model to transcribe audio into preliminary hypotheses or N-best lists.
These hypotheses serve as queries to match against the keyword dictionary.
Existing methods predominantly rely on dense text embeddings to perform this matching.
For instance, \citet{mathur-etal-2024-doc} utilize BERT \cite{devlin-etal-2019-bert}, while \citet{xiao-etal-2025-contextual} employ MPNet \cite{song-etal-2020-mpnet} to encode both hypotheses and keywords, ranking candidates based on semantic similarity.
The advantage of this paradigm lies in its training-free nature, allowing for the direct application of powerful text embedding models to capture rich semantic knowledge.

While effective for capturing semantic similarity, these semantic-only retrievers are susceptible to phonetic recognition errors, which are prevalent in Chinese.
Since base ASR models frequently produce homophonic or near-homophonic errors, the transcribed text often diverges significantly in meaning despite sounding similar to the target.
Consequently, retrievers relying solely on semantic embeddings are misled by these textual errors, fetching keywords that are semantically related to the incorrect transcripts but irrelevant to the actual speech.
Such irrelevant candidates fail to provide accurate context, rendering them ineffective for guiding the downstream contextual ASR model.
This limitation underscores that relying solely on global semantics is insufficient, necessitating the incorporation of other features, such as phonetic and glyph features.

\begin{figure*}[t]
    \centering
    \includegraphics[width=0.8\linewidth]{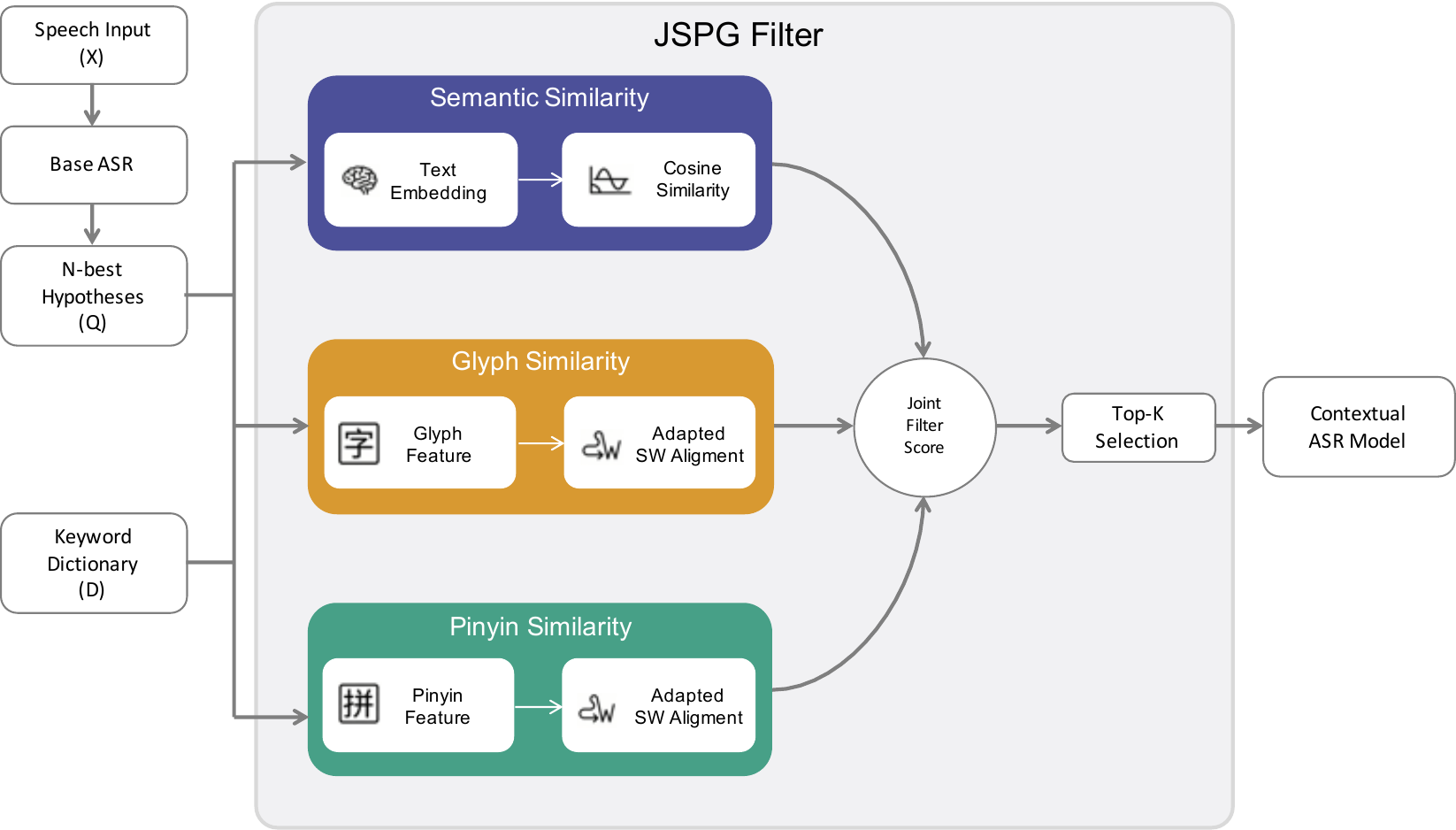}
    \caption{Overview of the proposed JSPG filtering framework. Given an input utterance, a base ASR model generates N-best hypotheses, which are compared to each candidate keyword in the dictionary via semantic, pinyin, and glyph similarity. The scores are combined to rank and retrieve the most relevant keywords for downstream contextual ASR.}
    \label{fig:overview}
\end{figure*}

\subsection{Audio-based End-to-End Retrieval}
In text-based cascaded methods, errors in this preliminary transcription will propagate to the retrieval stage.
It may lead to the failure of retrieving the correct keywords.
To avoid the reliance on intermediate text, audio-based end-to-end retrieval directly utilizes the input speech to retrieve keywords.
Specifically, these methods employ a speech encoder to generate embeddings for the audio and a text encoder to generate embeddings for the keywords.
They are trained using a contrastive loss to maximize the similarity score between the input speech and the relevant ground-truth keywords \cite{li2024rag, gong2025br}.
While this approach effectively bypasses error propagation, it typically necessitates extensive additional training on large-scale paired datasets to align the audio and text representations, making it data-hungry and computationally expensive.

In contrast, our proposed JSPG framework maintains the cascaded paradigm to directly leverage the sophisticated semantic capabilities of LLMs without additional training.
Crucially, we mitigate the negative impact of transcription errors by integrating phonetic and glyph features.

\subsection{Glyph and Pinyin Features in Chinese}
Chinese characters are unique in that they integrate semantics, phonetics, and structures \cite{tan-etal-2005-neuroanatomical}.
Specifically, phonetic and structural attributes are commonly measured by pinyin and glyph features, respectively.
Previous works in Chinese Spelling Check (CSC) and Named Entity Recognition (NER) have demonstrated that integrating these features is essential for resolving ambiguity.

Recently, \citet{qiao-etal-2025-disc} proposed an approach for quantifying the similarity between Chinese characters based on their phonetic and structural attributes.
However, it is designed for \textit{character-level} comparison and cannot be directly applied to our task.
The core challenge lies in computing the similarity between the keyword and the hypothesis sequence.
Typically, there is a significant length discrepancy between the short keyword and the longer hypothesis, and the hypothesis sequence usually contains recognition errors.
To bridge this gap, our JSPG method adapts the character-level similarity scores into a modified Smith-Waterman algorithm \cite{smith-waterman-1981-identification}.
This allows us to leverage phonetic and glyph cues to identify the optimal matching part within the hypothesis and return a sequence-level similarity score, which is used to assist the retrieval process in JSPG.

%% file: contents/approach.tex
\section{Our JSPG Approach}

\paragraph{Big Picture.} 
Our goal is to determine a small number of keywords that are really related to the input speech from a keyword dictionary containing numerous keywords. Only the selected few keywords are provided to the subsequent contextual ASR models. 
Specifically, we follow the N-best hypothesis paradigm \cite{mathur-etal-2024-doc, dimitrios-etal-2025-retrieval}, which consists of two steps. 
First, we obtain N-best transcriptions using a non-contextual basic ASR model. Second, we obtain the $K$ most related keywords from the dictionary using the N transcriptions. 
Previous works use implicit embedding-based similarity to represent relatedness, using LLMs to encode each transcription and each keyword into dense vectors. 
The main contribution of this work is incorporating explicit character similarity based on pronunciation and glyph. 
We follow the recent work on Chinese spelling check of \cite{qiao-etal-2025-disc}, and borrow the method for computing character similarity. 
To compute the relatedness between a transcription and a keyword, we extend the standard Smith-Waterman  algorithm \cite{smith-waterman-1981-identification} to incorporate character similarity.

\subsection{Problem Formulation}
Given an input speech $x$ and a large dictionary $D = \{w_1, w_2, \dots, w_M\}$, the goal of our JSPG is to filter $D$ into a concise, high-relevance subset of a predefined size $K$ ($K \ll M$).
Following previous practices \cite{xiao-etal-2025-contextual}, the process begins with a base ASR model that transcribes the speech $x$ into a preliminary list of N-best hypotheses $Q = \{q_1, q_2, \dots, q_N\}$, where each $q_j$ is a character sequence representing a transcription.
JSPG computes a joint score $F(Q, w)$ between $w$ and $Q$.
The top-K keywords are selected as the subset.
Specifically, $F(Q, w)$ is computed by integrating a semantic score $F^s(Q, w)$ and a phonetic-glyph score $F^{pg}(Q, w)$.
The phonetic-glyph score fuses pinyin-based score $F^p(Q, w)$ and glyph-based score $F^g(Q, w)$.
The final score combines semantic and phonetic-glyph scores:
\begin{equation}
    \label{eq:pg_score}
    F^{pg}(Q, w_i) = \alpha \cdot F^{p}(Q, w_i) + (1-\alpha) \cdot F^{g}(Q, w_i).
\end{equation}
\begin{equation}
    \label{eq:final_score}
    F(Q, w_i) = \beta \cdot F^{s}(Q, w_i) + (1-\beta) \cdot F^{pg}(Q, w_i).
\end{equation}

\subsection{Semantic-based Sequence Similarity}

To calculate the semantic similarity between the ASR hypotheses and the candidate keywords, we leverage dense text embeddings. 
We begin by concatenating the N-best hypotheses into a single sequence $Q'$, following the approach in \cite{dimitrios-etal-2025-retrieval}. 
Next, we employ the Qwen3-Embedding model \cite{zhang-etal-2025-qwen3} to encode both $Q'$ and the candidate keywords $w_i$. 
To maximize the model's retrieval capability, we follow its recommended settings and add a task-specific instruction before $Q'$: ``\emph{Given a list of candidate transcriptions predicted by a speech recognition model as a query, retrieve keywords relevant to the query. The candidate transcriptions are: $q_1$, $q_2$, $\dots$, $q_N$.}''.
The semantic score $F^s(Q, w_i)$ is then derived using cosine similarity:
\begin{equation}
\label{eq:semantic_score}
F^{s}(Q, w_i) = \frac{\text{Emb}(Q') \cdot \text{Emb}(w_i)}{\|\text{Emb}(Q')\| \, \|\text{Emb}(w_i)\|}
\end{equation}

While text embeddings effectively capture the global context, ASR errors in $Q'$ often change the actual meaning of the transcription, causing pure semantic retrieval to fail. This limitation necessitates the incorporation of pinyin and glyph features.

\subsection{Phonetic-Glyph Character Similarity}
\label{sec:phonetic_structural}

ASR prediction errors typically disrupt the semantic context but preserve the phonetic cues of the target keyword. To retrieve the target keywords using this cue, we compute character-level similarity using two complementary features: pinyin and glyph. First, pinyin directly encodes pronunciation in Chinese, so it can be used to measure phonetic similarity and recall the target keywords. However, in Chinese, a single pronunciation often corresponds to multiple distinct characters. Therefore, using pinyin to retrieve keywords inevitably fetches many irrelevant candidates.
Second, the glyph feature serves as another effective metric to retrieve the target keywords. 
Because most Chinese characters are phono-semantic compounds, characters with similar pronunciations frequently share similar structural components. Furthermore, unlike the one-to-many mapping of pinyin, every character has a unique glyph structure. The one-to-one property enables the glyph feature to effectively exclude irrelevant keywords that sound similar to the target.

\paragraph{Pinyin Similarity.}
Following \citet{qiao-etal-2025-disc}, for any two characters $c_1$ and $c_2$, we extract their complete pinyin sequences, denoted as $\texttt{py}(c_1)$ and $\texttt{py}(c_2)$. The pinyin similarity score $sim_p(c_1, c_2)$ is then calculated using the normalized Levenshtein Distance (LD):
\begin{equation}
\label{eq:pinyin_sim}
sim_p(c_1, c_2) = 1 - \frac{\texttt{LD}(\texttt{py}(c_1), \texttt{py}(c_2))}{|\texttt{py}(c_1)| + |\texttt{py}(c_2)|},
\end{equation}
where $|\cdot|$ represents the length of the pinyin string.

\paragraph{Glyph Similarity.}
To measure the glyph similarity between characters, we adopt the character-level metric proposed by \citet{qiao-etal-2025-disc}. This metric comprehensively evaluates glyph similarity across four distinct dimensions. For any two characters $c_1$ and $c_2$, the glyph similarity score $sim_g(c_1, c_2)$ is defined as the average of four sub-metrics:
\begin{enumerate}
\item \textbf{Four-corner Code ($sim_g^1$)}: Measures the similarity between numerical codes that represent the shapes of the character's four corners.
\item \textbf{Structure-aware Code ($sim_g^2$)}: Encodes radical decomposition and spatial layout (e.g., left-right vs. top-bottom structures).
\item \textbf{Stroke Sequence Edit Distance ($sim_g^3$)}: Computes the normalized LD between standard stroke-order sequences to capture fine-grained writing differences.
\item \textbf{Stroke Sequence LCS ($sim_g^4$)}: Computes the normalized Longest Common Subsequence (LCS) of stroke sequences to identify shared internal writing patterns.
\end{enumerate}

The final character-level glyph similarity is the average of these four components:
\begin{equation}\label{eq:char_glyph_sim}sim_g(c_1, c_2) = \frac{1}{4}\sum_{k=1}^{4}sim_{g}^{k}(c_1, c_2).\end{equation}

Currently, both $sim_p$ and $sim_g$ operate strictly at the character level. However, a key challenge is extending them to quantify the sequence-level relevance between a short keyword and a longer, noisy ASR hypothesis sequence. To bridge this gap, we introduce an extended Smith-Waterman algorithm to extend these character-level metrics into sequence-level scores, yielding $F^p(Q, w_i)$ and $F^g(Q, w_i)$.

\subsection{Sequence-Level Similarity via an Extended Smith-Waterman Algorithm}
\label{sec:sw}

The similarities defined above operate at the single-character level. In other words, given a pair of characters, $sim_g(.)$ (or $sim_p(.)$) can give a similarity score between them.

However, our retrieval module needs to assess how likely it is that a short keyword sequence $\boldsymbol{w}=c_1c_2...c_s$ is related to a longer sequence $\boldsymbol{q}=x_1x_2...x_n$. 
Specifically, $\boldsymbol{w}$ is considered related to $\boldsymbol{q}$ if $\boldsymbol{w}$ itself, or a modified version of it, is a substring of $\boldsymbol{q}$.
To achieve this, we employ an extended Smith-Waterman algorithm \cite{smith-waterman-1981-identification}, a dynamic programming approach based on edit distance.

\paragraph{Edit Distance.}
As a preliminary, the edit distance algorithm uses dynamic programming to efficiently compute the distance between two sequences, such as $a_1a_2...a_s$ and $b_1b_2...b_t$. We use $\texttt{ED}(\cdot)$ to denote the edit distance. Specifically, $\texttt{ED}(a_1a_2...a_s, b_1b_2...b_t)$ represents the total weight of the operations required to transform $a_1a_2...a_s$ into $b_1b_2...b_t$. There are four standard operations: match (no change), deletion, insertion, and substitution. Their corresponding weights are typically set to 0, 1.0, 1.0, and 1.0, respectively.

\paragraph{Standard Smith-Waterman.}
The standard Smith-Waterman algorithm is built upon $\texttt{ED}(.)$ and a more complex dynamic programming algorithm. 
Given $\boldsymbol{w} = c_1...c_s$ and $\boldsymbol{q} = x_1...x_n$, the algorithm finds the optimal indices $u, v$ ($1 \le u \le v \le s$) for $\boldsymbol{w}$ and $i, j$ ($1 \le i \le j \le n$) for $\boldsymbol{q}$, which yield the lowest edit distance between the substrings $c_u...c_v$ and $x_i...x_j$.
$$\texttt{SW}(\boldsymbol{w}, \boldsymbol{q}) = \min_{u,v,i,j} \texttt{ED}(c_u...c_v, x_i...x_j)$$

\paragraph{Extended $\texttt{SW}(.)$.}
To accommodate our situation, we make a two-fold extension. First, we further consider character similarity for the substitution operation. Taking $sim_{g}$ as an example, we change the weight of substituting $c_1$ with $c_2$ from a fixed value of $1.0$ to:
$$ 1 - sim_{g}(c_1, c_2)$$
The more similar $c_1$ is to $c_2$, the lower the weight. When $c_1 = c_2$, which corresponds to the not-change operation, the weight is zero.

Second, we prohibit deletion operations at the beginning and end positions of $\boldsymbol{w}$ (i.e., for $c_1$ and $c_s$). This constraint mathematically enforces $u=1$ and $v=s$, forcing the entirety of $\boldsymbol{w}$ to be aligned against a substring of $\boldsymbol{q}$.

Then, the relatedness likelihood is computed as:
$$\texttt{RL}(\boldsymbol{w}, \boldsymbol{q}) = \frac{|\boldsymbol{w}|-\texttt{SW}_{\text{ext}}(\boldsymbol{w}, \boldsymbol{q})}{|\boldsymbol{w}|}$$

Considering that the base ASR model typically generates an N-best list of hypotheses $Q=\{q_{1},q_{2},...,q_{N}\}$  rather than a single transcription, we compute the relatedness likelihood $\texttt{RL}(w, q_{j})$ for each individual hypothesis $q_{j} \in Q$. 
We take the maximum likelihood score across all N hypotheses. Therefore, the final sequence-level similarity score between the keyword $w$ and the N-best hypotheses $Q$ is formulated as:
$$F^{*}(Q, w) = \max_{q_{j} \in Q} \texttt{RL}(w, q_{j})$$
where $*$ represents either $p$ or $g$, depending on whether pinyin or glyph features are used when calculating the relatedness likelihood \texttt{RL}.

\input{figs/retriever}

%% file: figs/retriever.tex
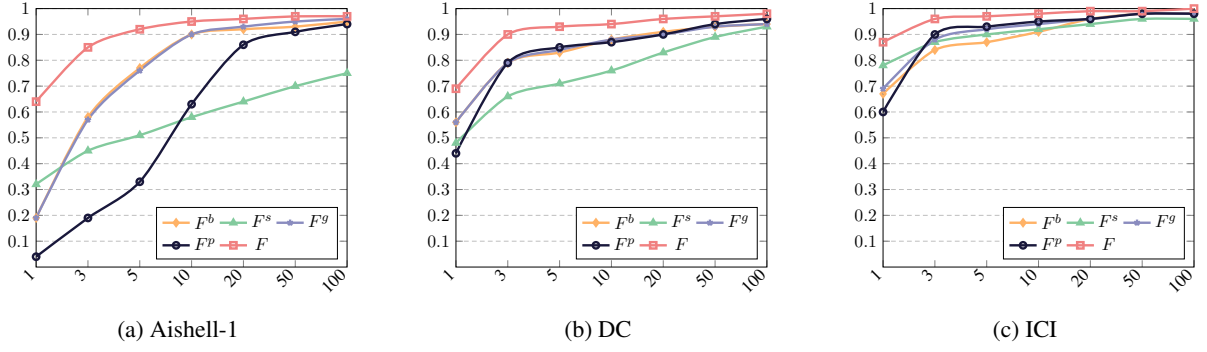
\begin{figure*}[!tb]
    \centering
    \begin{subfigure}[b]{0.3\linewidth}
      \centering
      \scalebox{0.6}{%
      \begin{tikzpicture}
        \begin{axis}[
            xmin=1,xmax=7,
            ymin=0.0,ymax=1.0,
            xtick={1, 2, 3, 4, 5, 6, 7},
            xticklabels={1, 3, 5, 10, 20, 50, 100}, %
            x tick label style={rotate=45,anchor=east},
            ytick={0.1, 0.2, 0.3, 0.4, 0.5, 0.6, 0.7, 0.8, 0.9, 1.0},
            ymajorgrids=true,
            major grid style={line width=.2pt,draw=gray!50,densely dashed},
            minor grid style={line width=.1pt,draw=gray!30},
            legend style={
                legend pos=south east,  %
                legend columns=3,       %
            },
            ]
            \addplot[color=myorange, line width=1.5pt, mark=diamond, smooth] coordinates {
              (1, 0.19) (2, 0.58) (3, 0.77) (4, 0.90) (5, 0.92) (6, 0.93) (7, 0.95)
            };
            \addlegendentry{$F^b$}
            \addplot[color=mygreen, line width=1.5pt, mark=triangle, smooth] coordinates {
              (1, 0.32) (2, 0.45) (3, 0.51) (4, 0.58) (5, 0.64) (6, 0.70) (7, 0.75)
            };
            \addlegendentry{$F^s$}
            \addplot[color=clearpurple, line width=1.5pt, mark=star, smooth] coordinates {
              (1, 0.19) (2, 0.57) (3, 0.76) (4, 0.90) (5, 0.93) (6, 0.95) (7, 0.96)
            };
            \addlegendentry{$F^{g}$}
            \addplot[color=darkblue, line width=1.5pt, mark=o, smooth] coordinates {
              (1, 0.04) (2, 0.19) (3, 0.33) (4, 0.63) (5, 0.86) (6, 0.91) (7, 0.94)
            };
            \addlegendentry{$F^{p}$}
            \addplot[color=myred, line width=1.5pt, mark=square, smooth] coordinates {
              (1, 0.64) (2, 0.85) (3, 0.92) (4, 0.95) (5, 0.96) (6, 0.97) (7, 0.97)
            };
            \addlegendentry{$F$}
        \end{axis}
      \end{tikzpicture}%
      }
      \caption{Aishell-1}
      \label{fig:retriever_aishell}
    \end{subfigure}%
    \hfill
    \begin{subfigure}[b]{0.3\linewidth}
      \centering
      \scalebox{0.6}{%
      \begin{tikzpicture}
        \begin{axis}[
            xmin=1,xmax=7,
            ymin=0.0,ymax=1.0,
            xtick={1, 2, 3, 4, 5, 6, 7},
            xticklabels={1, 3, 5, 10, 20, 50, 100},
            x tick label style={rotate=45,anchor=east},
            ytick={0.1, 0.2, 0.3, 0.4, 0.5, 0.6, 0.7, 0.8, 0.9, 1.0},
            ymajorgrids=true,
            major grid style={line width=.2pt,draw=gray!50,densely dashed},
            minor grid style={line width=.1pt,draw=gray!30},
            legend style={
                legend pos=south east,  %
                legend columns=3,       %
            },
            ]
            \addplot[color=myorange, line width=1.5pt, mark=diamond, smooth] coordinates {
              (1, 0.56) (2, 0.79) (3, 0.83) (4, 0.88) (5, 0.91) (6, 0.93) (7, 0.94)
            };
            \addlegendentry{$F^b$}
            \addplot[color=mygreen, line width=1.5pt, mark=triangle, smooth] coordinates {
              (1, 0.48) (2, 0.66) (3, 0.71) (4, 0.76) (5, 0.83) (6, 0.89) (7, 0.93)
            };
            \addlegendentry{$F^s$}
            \addplot[color=clearpurple, line width=1.5pt, mark=star, smooth] coordinates {
              (1, 0.56) (2, 0.79) (3, 0.84) (4, 0.88) (5, 0.90) (6, 0.93) (7, 0.94)
            };
            \addlegendentry{$F^{g}$}
            \addplot[color=darkblue, line width=1.5pt, mark=o, smooth] coordinates {
              (1, 0.44) (2, 0.79) (3, 0.85) (4, 0.87) (5, 0.90) (6, 0.94) (7, 0.96)
            };
            \addlegendentry{$F^{p}$}
            \addplot[color=myred, line width=1.5pt, mark=square, smooth] coordinates {
              (1, 0.69) (2, 0.90) (3, 0.93) (4, 0.94) (5, 0.96) (6, 0.97) (7, 0.98)
            };
            \addlegendentry{$F$}
        \end{axis}
      \end{tikzpicture}%
      }
      \caption{DC}
      \label{fig:retriever_dc}
    \end{subfigure}
    \hfill
    \begin{subfigure}[b]{0.3\linewidth}
      \centering
      \scalebox{0.6}{%
      \begin{tikzpicture}
        \begin{axis}[
            xmin=1,xmax=7,
            ymin=0.0,ymax=1.0,
            xtick={1, 2, 3, 4, 5, 6, 7},
            xticklabels={1, 3, 5, 10, 20, 50, 100},
            x tick label style={rotate=45,anchor=east},
            ytick={0.1, 0.2, 0.3, 0.4, 0.5, 0.6, 0.7, 0.8, 0.9, 1.0},
            ymajorgrids=true,
            major grid style={line width=.2pt,draw=gray!50,densely dashed},
            minor grid style={line width=.1pt,draw=gray!30},
            legend style={
                legend pos=south east,  %
                legend columns=3,       %
            },
            ]
            \addplot[color=myorange, line width=1.5pt, mark=diamond, smooth] coordinates {
              (1, 0.67) (2, 0.84) (3, 0.87) (4, 0.91) (5, 0.96) (6, 0.98) (7, 0.98)
            };
            \addlegendentry{$F^b$}
            \addplot[color=mygreen, line width=1.5pt, mark=triangle, smooth] coordinates {
              (1, 0.78) (2, 0.87) (3, 0.90) (4, 0.92) (5, 0.94) (6, 0.96) (7, 0.96)
            };
            \addlegendentry{$F^s$}
            \addplot[color=clearpurple, line width=1.5pt, mark=star, smooth] coordinates {
              (1, 0.69) (2, 0.88) (3, 0.92) (4, 0.94) (5, 0.96) (6, 0.98) (7, 0.98)
            };
            \addlegendentry{$F^{g}$}
            \addplot[color=darkblue, line width=1.5pt, mark=o, smooth] coordinates {
              (1, 0.60) (2, 0.90) (3, 0.93) (4, 0.95) (5, 0.96) (6, 0.98) (7, 0.98)
            };
            \addlegendentry{$F^{p}$}
            \addplot[color=myred, line width=1.5pt, mark=square, smooth] coordinates {
              (1, 0.87) (2, 0.96) (3, 0.97) (4, 0.98) (5, 0.99) (6, 0.99) (7, 1.00)
            };
            \addlegendentry{$F$}
        \end{axis}
      \end{tikzpicture}%
      }
      \caption{ICI}
      \label{fig:retriever_ici}
    \end{subfigure}
  \caption{The performance of retrieval methods on Aishell, DC, and ICI datasets. The x-axis indicates the K value in top-K, and the y-axis shows the corresponding Recall@K.}
    \label{fig:retriever_keyword_list_effect}
  \end{figure*}

%% file: contents/exp.tex
\section{Experiments}
\subsection{Experimental Setup}

\paragraph{Datasets.}
To comprehensively evaluate the proposed approach across varying domains and speaking styles, we conduct experiments on two datasets. Aishell-1 \cite{bu-2017-aishell,chen-2022-aishell} serves as our standard benchmark, containing 150 hours of Mandarin speech recorded in a clean environment. 
To validate robustness in complex, real-world scenarios, we utilize the RWCS-NER dataset \cite{zhou-etal-2024-chinese}, which comprises two distinct domains: open-domain daily conversation (DC) and intelligent cockpit instructions (ICI). It features spontaneous speech, background noise, and a high density of rare entities, making it a challenging task for contextual ASR.

\paragraph{Pipeline Settings.}
We evaluate our approach using a three-stage pipeline. 
First, we employ Whisper \cite{radford-2022-whisper}, fine-tuned on the Aishell-1 training set, as the base ASR model to generate the initial N-best hypotheses $Q$. 
Second, for the retrieval stage, we compare our JSPG against an exact character-matching baseline ($F^b$) that uses the same Smith-Waterman framework, as well as single-feature retrievers relying solely on semantic ($F^s$, powered by Qwen3-Embedding-0.6B \cite{zhang-etal-2025-qwen3}), glyph ($F^g$), or pinyin ($F^p$) scores. 
Finally, to verify downstream effectiveness, we integrate the retrieved keywords into two distinct contextual ASR paradigms: end-to-end models (CopyNE \cite{zhou-etal-2024-copyne} and a multi-grained model \cite{zhou2025improving}) and a two-stage LLM-refinement paradigm utilizing GPT-OSS-120B \cite{openai2025gptoss120bgptoss20bmodel} to correct the N-best hypotheses based on the retrieved prompts.

\paragraph{Evaluation Metrics}
We evaluate our system from two perspectives: the quality of the retrieved keywords and the final ASR recognition accuracy.
To assess retrieval quality, we adopt Recall@$K$ (R@$K$) as our core metric \cite{chen-etal-2025-wavrag}. It measures the proportion of ground-truth keywords captured within the top-K retrieved list.
In our experiments, we report R@$K$ for $K \in \{1, 3, 5, 10, 20, 50, 100\}$.

To assess downstream recognition accuracy and verify whether the retrieved keywords effectively guide the ASR model, we report four metrics. We measure the overall Character Error Rate (\textbf{CER}) on the full text, and the Unbiased-CER (\textbf{U-CER}) on non-keyword text to ensure general transcription quality is preserved. For keyword-specific performance, we report Keyword-CER (\textbf{K-CER}) on the keyword segments and Keyword Recall (\textbf{Recall}), which represents the percentage of ground-truth keywords correctly transcribed.

\subsection{Results of Retrieval}

\paragraph{Overall Trend.}
Figure \ref{fig:retriever_keyword_list_effect} illustrates the recall performance as $K$ varies. A consistent trend is observed across all datasets: the recall of all retrieval methods increases monotonically with $K$ and eventually saturates. This aligns with the theoretical expectation that a larger candidate pool covers more ground-truth targets. Crucially, our proposed JSPG algorithm ($F$) consistently achieves the highest recall among all methods across all $K$ settings. This empirical evidence validates that fusing semantic, glyph, and pinyin features provides a more robust retrieval signal than any single modality or the character-matching baseline.

\paragraph{Comparison between Different Retrieval Approaches.}
The performance gap between methods is most pronounced at lower $K$ values (e.g., $K=1, 3$), which reflect the precision of the retriever. In this setting, the joint algorithm $F$ demonstrates a decisive advantage, significantly outperforming other methods. 
Notably, on the relatively clean Aishell-1 and DC datasets, the character-matching baseline $F^b$ performs comparably to the glyph-based method $F^g$. This suggests that many keywords in these scenarios can be recalled via surface-level or visual similarity. 
Conversely, while the pinyin-based retriever $F^p$ underperforms at $K=1$ (especially on Aishell, dropping to 0.04), its recall rate grows rapidly as $K$ increases, eventually matching or surpassing $F^b$ and $F^g$ (e.g., reaching 0.96 on DC at $K=100$). This indicates that pinyin features are highly effective at capturing phonetic confusion and homophonic errors, provided a slightly larger candidate pool is allowed to accommodate phonetic ambiguity.

\input{tables/ragasr-overall}

\input{tables/weight_ablation.tex}
\paragraph{Ablation on Phonetic-Glyph Feature Weights.}
\label{sec:ablation_pg}
We conduct a two-stage ablation study to investigate the complementarity among semantic, pinyin, and glyph features, as summarized in Table \ref{tab:ablation_compact}.

\textit{Stage 1: Pinyin vs. Glyph ($\alpha$)}
Varying the pinyin weight $\alpha$ reveals a strong symbiotic relationship. 
Interestingly, R@1 remains highly stable ($\sim$18.7\%) across all settings. This stability demonstrates that top-1 precision is heavily anchored by the glyph feature, whose one-to-one structural mapping successfully filters out exact homophone confusions. However, for a larger candidate pool, R@100 rises substantially to peak at 99.13\% when $\alpha=0.7$. This proves that the pinyin feature is indispensable for capturing a broader range of structurally distinct but phonetically similar ASR errors. We set $\alpha=0.7$ to optimally balance broad phonetic tolerance and precise structural filtering.

\textit{Stage 2: Semantic vs. Phonetic-Glyph ($\beta$).}
Fixing $\alpha=0.7$, we evaluate the semantic weight $\beta$ (Table \ref{tab:ablation_compact}) to demonstrate the necessity of feature fusion.
As shown in Figure \ref{fig:retriever_keyword_list_effect}, the recall curve for pure semantic retrieval ($F^s$) flattens out early. 
Because homophonic ASR errors severely distort sentence meaning, pure semantics assigns extremely low scores to the correct targets, pushing them completely out of the top-100 pool. Table \ref{tab:ablation_compact} confirms this limitation: over-relying on semantics ($\beta=0.9$) causes the overall recall to drop to 89.11\%.
In contrast, fusing semantic and phonetic-glyph features effectively overcomes this bottleneck. At the optimal weight ($\beta=0.4$), the system achieves R@1 = 64.52\% while sustaining a high overall recall (R@100 = 97.03\%). 
This confirms that the semantic and phonetic-glyph features are highly complementary, and their fusion leads to significantly better retrieval performance.

\subsection{Results of Contextual ASR}
Having validated the retrieval performance, we now evaluate the impact of the proposed JSPG retriever on the downstream contextual ASR task. 
We integrate JSPG into the pipeline and test it with three different recognition models: CopyNE (Deep Biasing), Multi-grained (LLM-Fusion), and LLM-Refine (Two-stage correction).
Table \ref{table:ragasr-overall} details the performance of the retrieval-augmented systems, with the zero-shot performance of the fine-tuned Whisper baseline provided directly in the dataset headers for direct comparison.

\paragraph{Improvements over Baseline}
Experimental results demonstrate that our framework significantly outperforms the Whisper baseline across all datasets. 
On the standard Aishell-1, the best-performing configuration reduces the K-CER from 10.40\% to 2.70\% and improves R from 80.60\% to 95.53\% (Multi-grained with Top-K=10).
The gains are even more pronounced in the real-world datasets. On the noisy, task-oriented ICI dataset, K-CER drops dramatically from 30.70\% to 4.85\%, and R surges from 40.80\% to 93.20\% (LLM-Refine with Top-K=10). 
These results confirm that incorporating retrieved context via JSPG effectively mitigates the keyword recognition errors.

\paragraph{Performance of Different Recognition Models}
The effectiveness of the recognition models varies by domain. 
The Multi-grained model performs best on Aishell-1 dataset. 
However, for the highly complex ICI dataset, the LLM-Refine method (utilizing GPT-OSS-120B) achieves the lowest K-CER. 
This indicates that while specialized end-to-end models are efficient for general tasks, large-scale LLMs possess stronger capabilities for keyword correction in difficult scenarios.
Importantly, the U-CER remains stable across all experiments, ensuring that our retrieval-based biasing does not negatively impact the recognition of general non-keyword text.

%% file: tables/ragasr-overall.tex
\begin{table*}[!tb]
    \centering
    \setlength{\tabcolsep}{1.5pt}
    \begin{tabular}{l cccc cccc cccc}
        \toprule
        & \multicolumn{4}{c}{CopyNE} & \multicolumn{4}{c}{Multi-grained} & \multicolumn{4}{c}{LLM-Refine} \\
        \cmidrule(lr){2-5} \cmidrule(lr){6-9} \cmidrule(lr){10-13}
        Top-K & CER$\downarrow$ & K-CER$\downarrow$ & U-CER$\downarrow$ & R$\uparrow$ & CER$\downarrow$ & K-CER$\downarrow$ & U-CER$\downarrow$ & R$\uparrow$ & CER$\downarrow$ & K-CER$\downarrow$ & U-CER$\downarrow$ & R$\uparrow$ \\
        \midrule

        \multicolumn{13}{c}{\cellcolor{gray!25} \textbf{Aishell-1} \quad {\small \textit{(Whisper Base \ $\vert$ \ CER: 5.20 \ $\vert$ \ K-CER: 10.40 \ $\vert$ \ U-CER: 4.70 \ $\vert$ \ R: 80.60)}}} \\
        1 & 4.80 & 5.60 & \textbf{4.72} & 88.19 & 3.97 & 4.90 & \textbf{3.87} & 89.43 & 4.45 & 7.00 & \textbf{4.18} & 85.77 \\
        10 & \textbf{4.70} & \textbf{3.90} & 4.78 & \textbf{93.51} & \textbf{3.84} & \textbf{2.70} & 3.96 & \textbf{95.53} & \textbf{4.16} & \textbf{3.30} & 4.25 & \textbf{93.23} \\
        50 & 4.74 & 4.20 & 4.80 & 92.80 & 3.86 & 3.10 & 3.94 & 94.42 & 4.19 & 3.90 & 4.22 & 91.55 \\
        100 & 4.78 & 4.35 & 4.82 & 92.53 & 3.90 & 3.35 & 3.96 & 94.06 & 4.27 & 4.40 & 4.26 & 90.80 \\

        \midrule
        \multicolumn{13}{c}{\cellcolor{gray!25} \textbf{DC} \quad {\small \textit{(Whisper Base \ $\vert$ \ CER: 12.80 \ $\vert$ \ K-CER: 22.90 \ $\vert$ \ U-CER: 11.70 \ $\vert$ \ R: 71.10)}}} \\
        1 & 12.25 & 18.90 & \textbf{11.65} & 76.62 & 11.40 & 14.76 & 11.09 & 82.05 & 11.51 & 14.50 & 11.24 & 81.56 \\
        10 & \textbf{12.10} & \textbf{15.55} & 11.79 & \textbf{81.33} & \textbf{10.94} & \textbf{11.90} & \textbf{10.85} & \textbf{86.02} & \textbf{11.13} & \textbf{12.00} & \textbf{11.05} & \textbf{86.17} \\
        50 & 12.18 & 15.90 & 11.84 & 80.80 & 11.04 & 12.30 & 10.92 & 85.25 & 11.32 & 12.70 & 11.19 & 85.09 \\
        100 & 12.26 & 16.10 & 11.91 & 80.44 & 11.22 & 12.90 & 11.07 & 84.80 & 11.36 & 13.10 & 11.20 & 84.98 \\

        \midrule
        \multicolumn{13}{c}{\cellcolor{gray!25} \textbf{ICI} \quad {\small \textit{(Whisper Base \ $\vert$ \ CER: 11.50 \ $\vert$ \ K-CER: 30.70 \ $\vert$ \ U-CER: 6.90 \ $\vert$ \ R: 40.80)}}} \\
        1 & 9.30 & 19.40 & \textbf{6.95} & 64.91 & 7.90 & 12.70 & \textbf{6.78} & 76.68 & 6.96 & 8.40 & 6.63 & 87.22 \\
        10 & \textbf{8.96} & \textbf{17.30} & 7.02 & \textbf{69.12} & \textbf{7.71} & \textbf{11.40} & 6.85 & \textbf{79.19} & \textbf{5.78} & \textbf{4.85} & 5.99 & \textbf{93.20} \\
        50 & 9.06 & 17.70 & 7.05 & 68.47 & 7.83 & 12.15 & 6.82 & 77.56 & 5.91 & 5.95 & \textbf{5.90} & 90.58 \\
        100 & 9.13 & 17.95 & 7.08 & 68.14 & 8.03 & 12.95 & 6.88 & 76.10 & 5.94 & 6.05 & 5.92 & 90.58 \\
        \bottomrule
    \end{tabular}
    \caption{Results of different contextual ASR models with JSPG retrieval. The performance of base ASR is provided in the dataset headers for direct comparison.}
    \label{table:ragasr-overall}
\end{table*}

%% file: tables/weight_ablation.tex
\begin{table}[!tb]
    \centering
    \small %
    \begin{tabular}{l c c c}
      \toprule
      \textbf{Ablation Phase} & \textbf{Weight} & \textbf{R@1} & \textbf{R@100} \\
      \midrule
      \multirow{3}{*}{\shortstack[l]{\textbf{Stage 1: $\alpha$}}} 
      & 0.1  & \textbf{18.72} & 96.50 \\
      & 0.7    & 18.68 & \textbf{99.13} \\
      & 0.9 & 18.68 & 98.82 \\
      \midrule
      \multirow{3}{*}{\shortstack[l]{\textbf{Stage 2: $\beta$}}} 
      & 0.1  & 62.31 & 96.94 \\
      & 0.4    & \textbf{64.52} & \textbf{97.03} \\
      & 0.9 & 43.16 & 89.11 \\
      \bottomrule
    \end{tabular}
    \caption{Ablation on feature weights. It reveals their complementary roles: glyph heavily anchors top-1 precision, pinyin expands the top-100 recall ceiling, and semantics acts as a vital tie-breaker upon a solid phonetic-glyph foundation.}
    \label{tab:ablation_compact}
  \end{table}

%% file: contents/conclusion.tex
\section{Conclusion}
In this paper, we introduced JSPG, a novel filtering framework designed to address the challenge of utilizing large-scale contextual dictionaries in ASR.
By synergizing semantic, glyph, and pinyin information, JSPG effectively bridges the gap between noisy hypotheses and target keywords, ensuring robust recall even in the presence of severe recognition errors.
We believe this work offers a practical solution for deploying contextual ASR in open-domain scenarios and paves the way for future research in speech processing.

%% file: contents/sw_example.tex
\subsection{Algorithmic Walkthrough of Extended Smith-Waterman}
\label{app:sw_example}

To demonstrate the process of our extended SW algorithm, we present a step-by-step walkthrough here.
This example computes the sequence-level similarity score between the keyword $w=$ ``语音识别'' (speech recognition, $|w|=4$) and the ASR hypothesis ``关于雨音的识别'' (about the recognition of rain sound).
Here the target character ``语'' is misrecognized as the homophone ``雨'' and an extra character ``的'' is inserted.

\paragraph{Input: Substitution Cost Matrix}
Table \ref{tab:cost_matrix} presents the substitution costs between the hypothesis characters and keyword characters. As defined in Section \ref{sec:sw}, each cost is computed as $1 - sim_p(c_i, c_j)$, where $sim_p$ is the pinyin similarity defined in Section \ref{sec:phonetic_structural}.
A cost of 0 indicates identical pronunciation. Notably, the homophonic pair (``雨'', ``语'') shares identical pinyin (\texttt{yu3}), resulting in a substitution cost of 0.

\paragraph{Process: Cost Accumulation Matrix ($D$)}
Using the substitution costs from Table \ref{tab:cost_matrix} and a gap cost of 1, we compute the cost matrix $D$, where $D_{i,j}$ represents the minimum alignment cost of the first $j$ keyword characters against hypothesis characters ending at position $i$. The initialization is $D_{i,0} = 0$ (free start in the hypothesis) and $D_{0,j} = \infty$ for $j \ge 1$.
For general positions, the recurrence is $D_{i,j} = \min(D_{i-1,j-1} + \text{cost}_{i,j},\, D_{i-1,j} + 1,\, D_{i,j-1} + 1)$.
As described in Section \ref{sec:sw}, for the first and last keyword positions ($j=1$ and $j=|w|$), the deletion move $D_{i,j-1}+1$ is excluded to ensure the entire keyword is matched.
The resulting matrix is shown in Table \ref{tab:d_matrix}.
We highlight the calculation for the following critical nodes.

\begin{table}[bt]
\centering
\small
\setlength{\tabcolsep}{3pt}
\begin{tabular}{l|cccc}
\toprule
\textbf{cost} & $\text{语}_{1}(\texttt{yu3})$ & $\text{音}_{2}(\texttt{yin1})$ & $\text{识}_{3}(\texttt{shi2})$ & $\text{别}_{4}(\texttt{bie2})$ \\
\midrule
$\text{关}_{1}(\texttt{guan1})$ & 0.50 & 0.33 & 0.56 & 0.56 \\
$\text{于}_{2}(\texttt{yu2})$   & 0.17 & 0.43 & 0.43 & 0.43 \\
$\text{雨}_{3}(\texttt{yu3})$   & \textbf{0.00} & 0.43 & 0.57 & 0.57 \\
$\text{音}_{4}(\texttt{yin1})$  & 0.43 & \textbf{0.00} & 0.50 & 0.38 \\
$\text{的}_{5}(\texttt{de})$    & 0.60 & 0.67 & 0.67 & 0.50 \\
$\text{识}_{6}(\texttt{shi2})$  & 0.57 & 0.50 & \textbf{0.00} & 0.38 \\
$\text{别}_{7}(\texttt{bie2})$  & 0.57 & 0.38 & 0.38 & \textbf{0.00} \\
\bottomrule
\end{tabular}
\caption{Substitution costs ($1 - sim_p$) derived from pinyin similarity. A cost of 0 indicates identical pronunciation.}
\label{tab:cost_matrix}
\end{table}

\begin{table}[bt]
\centering
\small
\renewcommand{\arraystretch}{2.1}
\setlength{\tabcolsep}{6pt}

\begin{tabular}{l|ccccc}
\toprule
$D$ & $\emptyset_{0}$ & $\text{语}_{1}$ & $\text{音}_{2}$ & $\text{识}_{3}$ & $\text{别}_{4}$ \\
\midrule
$\emptyset_{0}$ & 0 & $\infty$ & $\infty$ & $\infty$ & $\infty$ \\
$\text{关}_{1}$ & 0 & 0.50 & 1.50 & 2.50 & $\infty$ \\
$\text{于}_{2}$ & \tikzmarknode{n20}{0} & 0.17 & 0.93 & 1.93 & 2.93 \\
$\text{雨}_{3}$ & 0 & \tikzmarknode{n31}{\textbf{0.00}} & 0.60 & 1.50 & 2.50 \\
$\text{音}_{4}$ & 0 & 0.43 & \tikzmarknode{n42}{\textbf{0.00}} & 1.00 & 1.88 \\
$\text{的}_{5}$ & 0 & 0.60 & \tikzmarknode{n52}{\textbf{1.00}} & 0.67 & 1.50 \\
$\text{识}_{6}$ & 0 & 0.57 & 1.10 & \tikzmarknode{n63}{\textbf{1.00}} & 1.04 \\
$\text{别}_{7}$ & 0 & 0.57 & 0.95 & 1.48 & \tikzmarknode{n74}{\textbf{1.00}} \\
\bottomrule
\end{tabular}

\begin{tikzpicture}[overlay, remember picture, >=Stealth, black, thick, shorten >= 2pt, shorten <= 2pt]
    \draw[->] (n20) -- (n31);
    \draw[->] (n31) -- (n42);
    \draw[->] (n42) -- (n52);
    \draw[->] (n52) -- (n63);
    \draw[->] (n63) -- (n74);
\end{tikzpicture}
\caption{The cost matrix $D$. Lower values indicate better alignment. The arrows visualize the optimal alignment path.}
\label{tab:d_matrix}
\end{table}

For $D_{3,1}$: Aligning $\text{雨}_{3}$ with $\text{语}_{1}$. Since $j=1$, deletion of the first keyword character is prohibited. With a substitution cost of $0$ (identical pinyin), the diagonal move yields the minimum:
\begin{equation*}
\begin{split}
    D_{3,1} &= \min(D_{2,0} + \text{cost}_{3,1},\, D_{2,1} + 1) \\
            &= \min(0 + 0,\, 0.17 + 1) = \mathbf{0}.
\end{split}
\end{equation*}

For $D_{4,2}$: Aligning $\text{音}_{4}$ with $\text{音}_{2}$. An exact match (cost $= 0$) continues the diagonal path:
\begin{equation*}
\begin{split}
    D_{4,2} &= \min(D_{3,1} + \text{cost}_{4,2},\, D_{3,2} + 1,\, D_{4,1} + 1) \\
            &= \min(0 + 0,\, 0.60 + 1,\, 0.43 + 1) = \mathbf{0}.
\end{split}
\end{equation*}

For $D_{5,2}$: The insertion character $\text{的}_{5}$ has a high substitution cost with $\text{音}_{2}$ (cost $= 0.67$).
The algorithm selects the vertical move $D_{4,2} + 1 = 1$, which inserts $\text{的}$ as a gap. This effectively extends the alignment of ``雨音'' $\leftrightarrow$ ``语音'' to handle the extra character, rather than breaking the existing match:
\begin{equation*}
\begin{split}
    D_{5,2} &= \min(D_{4,1} + \text{cost}_{5,2},\, D_{4,2} + 1,\, D_{5,1} + 1) \\
            &= \min(1.10,\, 1.00,\, 1.60) = \mathbf{1.00}.
\end{split}
\end{equation*}

For $D_{6,3}$ and $D_{7,4}$: After bypassing the insertion, exact matches (cost $= 0$) resume. Since $j=4$ at the last step, deletion is again prohibited:
\begin{align*}
    D_{6,3} &= \min(D_{5,2} + \text{cost}_{6,3},\, D_{5,3} + 1,\, D_{6,2} + 1) \\
            &= \min(1.00,\, 1.67,\, 2.10) = \mathbf{1.00}, \\[5pt]
    D_{7,4} &= \min(D_{6,3} + \text{cost}_{7,4},\, D_{6,4} + 1) \\
            &= \min(1.00,\, 2.04) = \mathbf{1.00}.
\end{align*}

Consequently, the keyword ``语音识别'' is successfully matched to the hypothesis part ``雨音的识别'', with a total alignment cost of $\texttt{SW}_{\text{ext}} = D_{7,4} = 1.0$. The relatedness likelihood is then computed as $\texttt{RL} = (|w| - \texttt{SW}_{\text{ext}}) / |w| = (4 - 1) / 4 = 0.75$.

%% file: acl_latex.bbl
\begin{thebibliography}{28}
\providecommand{\natexlab}[1]{#1}

\bibitem[{Alon et~al.(2019)Alon, Pundak, and Sainath}]{alon2019contextual}
Uri Alon, Golan Pundak, and Tara~N Sainath. 2019.
\newblock Contextual speech recognition with difficult negative training examples.
\newblock In \emph{ICASSP 2019-2019 IEEE International Conference on Acoustics, Speech and Signal Processing (ICASSP)}, pages 6440--6444. IEEE.

\bibitem[{Bai et~al.(2024)Bai, Chen, Chen, Chen, Chen, Ding, Dong, Dong, Du, Gao et~al.}]{bai-etal-2024-seedasr}
Ye~Bai, Jingping Chen, Jitong Chen, Wei Chen, Zhuo Chen, Chuang Ding, Linhao Dong, Qianqian Dong, Yujiao Du, Kepan Gao, and 1 others. 2024.
\newblock Seed-asr: Understanding diverse speech and contexts with llm-based speech recognition.
\newblock \emph{arXiv preprint arXiv:2407.04675}.

\bibitem[{Bu et~al.(2017)Bu, Du, Na, Wu, and Zheng}]{bu-2017-aishell}
Hui Bu, Jiayu Du, Xingyu Na, Bengu Wu, and Hao Zheng. 2017.
\newblock Aishell-1: An open-source mandarin speech corpus and a speech recognition baseline.
\newblock In \emph{2017 20th Conference of the Oriental Chapter of the International Coordinating Committee on Speech Databases and Speech I/O Systems and Assessment (O-COCOSDA)}, pages 1--5.

\bibitem[{Chen et~al.(2022)Chen, Xu, Wang, Xie, Zhang, and Huang}]{chen-2022-aishell}
Boli Chen, Guangwei Xu, Xiaobin Wang, Pengjun Xie, Meishan Zhang, and Fei Huang. 2022.
\newblock Aishell-ner: Named entity recognition from chinese speech.
\newblock In \emph{2022 IEEE International Conference on Acoustics, Speech and Signal Processing (ICASSP)}, pages 8352--8356.

\bibitem[{Chen et~al.(2025)Chen, Ji, Wang, Wang, Chen, He, Xu, and Zhao}]{chen-etal-2025-wavrag}
Yifu Chen, Shengpeng Ji, Haoxiao Wang, Ziqing Wang, Siyu Chen, Jinzheng He, Jin Xu, and Zhou Zhao. 2025.
\newblock \href {https://doi.org/10.18653/v1/2025.acl-long.613} {{W}av{RAG}: Audio-integrated retrieval augmented generation for spoken dialogue models}.
\newblock In \emph{Proceedings of the 63rd Annual Meeting of the Association for Computational Linguistics (Volume 1: Long Papers)}, pages 12505--12523, Vienna, Austria. Association for Computational Linguistics.

\bibitem[{Devlin et~al.(2019)Devlin, Chang, Lee, and Toutanova}]{devlin-etal-2019-bert}
Jacob Devlin, Ming-Wei Chang, Kenton Lee, and Kristina Toutanova. 2019.
\newblock {BERT}: Pre-training of deep bidirectional transformers for language understanding.
\newblock In \emph{Proceedings of the Conference of the North American Chapter of the Association for Computational Linguistics (NAACL)}, pages 4171--4186, Minneapolis, Minnesota. Association for Computational Linguistics.

\bibitem[{Dimitrios et~al.(2025)Dimitrios, Papadopoulos, Parada, Zhang, Saravanan, and Drosou}]{dimitrios-etal-2025-retrieval}
Siskos Dimitrios, Stavros Papadopoulos, Pablo~Peso Parada, Jisi Zhang, Karthikeyan Saravanan, and Anastasios Drosou. 2025.
\newblock \href {https://doi.org/10.18653/v1/2025.findings-emnlp.768} {Retrieval augmented generation based context discovery for {ASR}}.
\newblock In \emph{Findings of the Association for Computational Linguistics: EMNLP 2025}, pages 14247--14254, Suzhou, China. Association for Computational Linguistics.

\bibitem[{Gong et~al.(2025)Gong, Lv, Wang, Zhu, and Qian}]{gong2025br}
Xun Gong, Anqi Lv, Zhiming Wang, Huijia Zhu, and Yanmin Qian. 2025.
\newblock Br-asr: Efficient and scalable bias retrieval framework for contextual biasing asr in speech llm.
\newblock \emph{arXiv preprint arXiv:2505.19179}.

\bibitem[{Lei et~al.(2025)Lei, Na, Xu, Pusateri, Van~Gysel, Zhang, Han, and Huang}]{lei-etal-2025-contextualization}
Zhihong Lei, Xingyu Na, Mingbin Xu, Ernest Pusateri, Christophe Van~Gysel, Yuanyuan Zhang, Shiyi Han, and Zhen Huang. 2025.
\newblock Contextualization of asr with llm using phonetic retrieval-based augmentation.
\newblock In \emph{ICASSP 2025-2025 IEEE International Conference on Acoustics, Speech and Signal Processing (ICASSP)}, pages 1--5. IEEE.

\bibitem[{Li et~al.(2024{\natexlab{a}})Li, Shang, Wei, Guo, Li, He, Zhang, and Yang}]{li-etal-2024-rag}
Shaojun Li, Hengchao Shang, Daimeng Wei, Jiaxin Guo, Zongyao Li, Xianghui He, Min Zhang, and Hao Yang. 2024{\natexlab{a}}.
\newblock La-rag: Enhancing llm-based asr accuracy with retrieval-augmented generation.
\newblock \emph{arXiv preprint arXiv:2409.08597}.

\bibitem[{Li et~al.(2024{\natexlab{b}})Li, Shang, Wei, Guo, Li, He, Zhang, and Yang}]{li2024rag}
Shaojun Li, Hengchao Shang, Daimeng Wei, Jiaxin Guo, Zongyao Li, Xianghui He, Min Zhang, and Hao Yang. 2024{\natexlab{b}}.
\newblock La-rag: Enhancing llm-based asr accuracy with retrieval-augmented generation.
\newblock \emph{arXiv preprint arXiv:2409.08597}.

\bibitem[{Mathur et~al.(2024)Mathur, Liu, Li, Ma, Karen, Ahmed, Manocha, and Zhang}]{mathur-etal-2024-doc}
Puneet Mathur, Zhe Liu, Ke~Li, Yingyi Ma, Gil Karen, Zeeshan Ahmed, Dinesh Manocha, and Xuedong Zhang. 2024.
\newblock \href {https://aclanthology.org/2024.lrec-main.457/} {{DOC}-{RAG}: {ASR} language model personalization with domain-distributed co-occurrence retrieval augmentation}.
\newblock In \emph{Proceedings of the 2024 Joint International Conference on Computational Linguistics, Language Resources and Evaluation (LREC-COLING 2024)}, pages 5132--5139, Torino, Italia. ELRA and ICCL.

\bibitem[{Omnilingual et~al.(2025)Omnilingual, Keren, Kozhevnikov, Meng, Ropers, Setzler, Wang, Adebara, Auli, Balioglu et~al.}]{omnilingual2025omnilingual}
ASR Omnilingual, Gil Keren, Artyom Kozhevnikov, Yen Meng, Christophe Ropers, Matthew Setzler, Skyler Wang, Ife Adebara, Michael Auli, Can Balioglu, and 1 others. 2025.
\newblock Omnilingual asr: Open-source multilingual speech recognition for 1600+ languages.
\newblock \emph{arXiv preprint arXiv:2511.09690}.

\bibitem[{OpenAI(2025)}]{openai2025gptoss120bgptoss20bmodel}
OpenAI. 2025.
\newblock \href {https://arxiv.org/abs/2508.10925} {gpt-oss-120b and gpt-oss-20b model card}.
\newblock \emph{Preprint}, arXiv:2508.10925.

\bibitem[{Peters et~al.(2018)Peters, Neumann, Iyyer, Gardner, Clark, Lee, and Zettlemoyer}]{peters2018deep}
Matthew~E. Peters, Mark Neumann, Mohit Iyyer, Matt Gardner, Christopher Clark, Kenton Lee, and Luke Zettlemoyer. 2018.
\newblock Deep contextualized word representations.
\newblock In \emph{Proceedings of the Conference of the North American Chapter of the Association for Computational Linguistics (NAACL)}, pages 2227--2237, New Orleans, Louisiana. Association for Computational Linguistics.

\bibitem[{Pundak et~al.(2018)Pundak, Sainath, Prabhavalkar, Kannan, and Zhao}]{pundak-2018-deep}
Golan Pundak, Tara~N Sainath, Rohit Prabhavalkar, Anjuli Kannan, and Ding Zhao. 2018.
\newblock Deep context: end-to-end contextual speech recognition.
\newblock In \emph{2018 IEEE spoken language technology workshop (SLT)}, pages 418--425. IEEE.

\bibitem[{Qiao et~al.(2025)Qiao, Zhou, Liu, Li, Zhang, Zhang, Li, Zhang, and Huang}]{qiao-etal-2025-disc}
Ziheng Qiao, Houquan Zhou, Yumeng Liu, Zhenghua Li, Min Zhang, Bo~Zhang, Chen Li, Ji~Zhang, and Fei Huang. 2025.
\newblock \href {https://doi.org/10.18653/v1/2025.acl-long.1373} {{DISC}: Plug-and-play decoding intervention with similarity of characters for {C}hinese spelling check}.
\newblock In \emph{Proceedings of the 63rd Annual Meeting of the Association for Computational Linguistics (Volume 1: Long Papers)}, pages 28312--28324, Vienna, Austria. Association for Computational Linguistics.

\bibitem[{Radford et~al.(2022)Radford, Kim, Xu, Brockman, McLeavey, and Sutskever}]{radford-2022-whisper}
Alec Radford, Jong~Wook Kim, Tao Xu, Greg Brockman, Christine McLeavey, and Ilya Sutskever. 2022.
\newblock Robust speech recognition via large-scale weak supervision.
\newblock \emph{arXiv preprint arXiv:2212.04356}.

\bibitem[{Smith et~al.(1981)Smith, Waterman et~al.}]{smith-waterman-1981-identification}
Temple~F Smith, Michael~S Waterman, and 1 others. 1981.
\newblock Identification of common molecular subsequences.
\newblock \emph{Journal of molecular biology}, 147(1):195--197.

\bibitem[{Song et~al.(2020)Song, Tan, Qin, Lu, and Liu}]{song-etal-2020-mpnet}
Kaitao Song, Xu~Tan, Tao Qin, Jianfeng Lu, and Tie-Yan Liu. 2020.
\newblock Mpnet: Masked and permuted pre-training for language understanding.
\newblock \emph{Advances in neural information processing systems}, 33:16857--16867.

\bibitem[{Sudo et~al.(2024{\natexlab{a}})Sudo, Fukumoto, Shakeel, Peng, and Watanabe}]{sudo-2024-contextualized}
Yui Sudo, Yosuke Fukumoto, Muhammad Shakeel, Yifan Peng, and Shinji Watanabe. 2024{\natexlab{a}}.
\newblock Contextualized automatic speech recognition with dynamic vocabulary.
\newblock In \emph{2024 IEEE Spoken Language Technology Workshop (SLT)}, pages 78--85. IEEE.

\bibitem[{Sudo et~al.(2024{\natexlab{b}})Sudo, Shakeel, Fukumoto, Peng, and Watanabe}]{sudo2024contextualized}
Yui Sudo, Muhammad Shakeel, Yosuke Fukumoto, Yifan Peng, and Shinji Watanabe. 2024{\natexlab{b}}.
\newblock Contextualized automatic speech recognition with attention-based bias phrase boosted beam search.
\newblock In \emph{ICASSP 2024-2024 IEEE International Conference on Acoustics, Speech and Signal Processing (ICASSP)}, pages 10896--10900. IEEE.

\bibitem[{Tan et~al.(2005)Tan, Laird, Li, and Fox}]{tan-etal-2005-neuroanatomical}
Li~Hai Tan, Angela~R Laird, Karl Li, and Peter~T Fox. 2005.
\newblock Neuroanatomical correlates of phonological processing of chinese characters and alphabetic words: A meta-analysis.
\newblock \emph{Human brain mapping}, 25(1):83--91.

\bibitem[{Xiao et~al.(2025)Xiao, Hou, Garcia-Romero, and Han}]{xiao-etal-2025-contextual}
Cihan Xiao, Zejiang Hou, Daniel Garcia-Romero, and Kyu~J Han. 2025.
\newblock Contextual asr with retrieval augmented large language model.
\newblock In \emph{ICASSP 2025-2025 IEEE International Conference on Acoustics, Speech and Signal Processing (ICASSP)}, pages 1--5. IEEE.

\bibitem[{Zhang et~al.(2025)Zhang, Li, Long, Zhang, Lin, Yang, Xie, Yang, Liu, Lin et~al.}]{zhang-etal-2025-qwen3}
Yanzhao Zhang, Mingxin Li, Dingkun Long, Xin Zhang, Huan Lin, Baosong Yang, Pengjun Xie, An~Yang, Dayiheng Liu, Junyang Lin, and 1 others. 2025.
\newblock Qwen3 embedding: Advancing text embedding and reranking through foundation models.
\newblock \emph{arXiv preprint arXiv:2506.05176}.

\bibitem[{Zhou and Li(2025)}]{zhou2025improving}
Shilin Zhou and Zhenghua Li. 2025.
\newblock Improving contextual asr via multi-grained fusion with large language models.
\newblock \emph{arXiv preprint arXiv:2507.12252}.

\bibitem[{Zhou et~al.(2024{\natexlab{a}})Zhou, Li, Gong, Zhang, Hong, and Zhang}]{zhou-etal-2024-chinese}
Shilin Zhou, Zhenghua Li, Chen Gong, Lei Zhang, Yu~Hong, and Min Zhang. 2024{\natexlab{a}}.
\newblock \href {https://doi.org/10.18653/v1/2024.findings-acl.111} {{C}hinese spoken named entity recognition in real-world scenarios: Dataset and approaches}.
\newblock In \emph{Findings of the Association for Computational Linguistics: ACL 2024}, pages 1872--1884, Bangkok, Thailand. Association for Computational Linguistics.

\bibitem[{Zhou et~al.(2024{\natexlab{b}})Zhou, Li, Hong, Zhang, Wang, and Huai}]{zhou-etal-2024-copyne}
Shilin Zhou, Zhenghua Li, Yu~Hong, Min Zhang, Zhefeng Wang, and Baoxing Huai. 2024{\natexlab{b}}.
\newblock \href {https://doi.org/10.18653/v1/2024.acl-long.147} {{C}opy{NE}: Better contextual {ASR} by copying named entities}.
\newblock In \emph{Proceedings of the 62nd Annual Meeting of the Association for Computational Linguistics (Volume 1: Long Papers)}, pages 2675--2686, Bangkok, Thailand. Association for Computational Linguistics.

\end{thebibliography}
